\documentclass[pdflatex,sn-mathphys-num]{sn-jnl}

\usepackage{graphicx}%
\usepackage{multirow}%
\usepackage{amsmath,amssymb,amsfonts}%
\usepackage{amsthm}%
\usepackage{mathrsfs}%
\usepackage[title]{appendix}%
\usepackage{xcolor}%
\usepackage{textcomp}%
\usepackage{manyfoot}%
\usepackage{booktabs}%
\usepackage{algorithm}%
\usepackage{algorithmicx}%
\usepackage{algpseudocode}%
\usepackage{listings}%

\theoremstyle{thmstyleone}%
\theoremstyle{thmstyletwo}%

\theoremstyle{thmstylethree}%

\begin{document}

\title[VR Spine Simulation]{Virtual-reality based patient-specific simulation of spine surgical procedures: A fast, highly automated and high-fidelity system for surgical education and planning}


\author*[1]{\fnm{Raj Kumar} \sur{Ranabhat}}\email{rajkumar.ranabhat@sunnybrook.ca}
\author[1,3]{\fnm{Tayler D} \sur{Ross}}\email{taylerdeclan.ross@mail.utoronto.ca}
\author[1]{\fnm{Tony} \sur{Jiao}}\email{t.jiao@mail.utoronto.ca}
\author[1,2,3]{\fnm{Jeremie} \sur{Larouche}}\email{jeremie.larouche@sunnybrook.ca}
\author[1,2,3]{\fnm{Joel} \sur{Finkelstein}}\email{joel.finkelstein@sunnybrook.ca}
\author[1,3]{\fnm{Michael} \sur{Hardisty}}\email{m.hardisty@utoronto.ca}

\affil*[1]{\orgname{Holland Bone and Joint Program, Sunnybrook Research Institute}, 
\orgaddress{\city{Toronto}, \country{Canada}}}

\affil[2]{\orgname{Division of Spine Surgery, Sunnybrook Health Sciences Centre}, 
\orgaddress{\city{Toronto}, \country{Canada}}}

\affil[3]{\orgname{Division of Orthopaedic Surgery, Department of Surgery, University of Toronto}, 
\orgaddress{\city{Toronto}, \country{Canada}}}




\abstract{\textbf{Purpose:}
Surgical training involves didactic teaching, mentor-led learning, surgical skills laboratories, and direct exposure to surgery; however, increasing clinical pressures have resulted in limited operating room (OR) exposure. This investigation leverages virtual reality (VR)  to provide a safe, convenient, and immersive interactive learning environment. VR training is typically based on standardized off-the-shelf scenarios that are not linked to a learner’s specific clinical challenges and are often underutilized within surgical training programs. This study aims to change this paradigm by making the simulation fit the learner by using Artificial Intelligence (AI) based computer vision methods to generate patient-specific simulations directly from Computed Tomography (CT) and Magnetic Resonance Imaging (MRI).  Learners can rehearse real cases they are managing as part of their education.

\textbf{Methods:}
This study focused on patient-specific spinal decompression simulation for spinal stenosis in a virtual OR. The objectives were: (1) automatic creation of 3D anatomical models from patient imaging, and (2) VR simulation of spinal decompression procedures, including laminectomy, disc resection, and foraminotomy. Model construction required multimodal fusion (registration) of CT and MRI, and segmentation of relevant anatomical structures. Segmentation performance was evaluated using the Dice Similarity Coefficient (DSC), and registration accuracy was quantified using Target Registration Error (TRE). Qualitative feedback was obtained from surgeons and trainees.

\textbf{Results:}
High-fidelity patient-specific 3D models were created quickly (approximately 2.5 minutes per case, N = 15). Segmentation accuracy was high, with a DSC of 0.95 ± 0.03 for vertebral bone and 0.895 ± 0.02 for soft tissue structures. Registration accuracy demonstrated a mean TRE of 1.73 ± 0.42 mm. Semi-structured interviews indicated improved spatial understanding, increased procedural confidence, and perceived educational value, with strong support for integration into clinical workflows.

\textbf{Conclusion:}
This platform significantly reduced the time and costs of patient-specific modelling, thereby facilitating pre-operative planning, post-procedural assessments, and comprehensive surgical simulation. By enhancing surgical training and pre-operative planning, this technology has the potential to improve patient outcomes, optimize operating room efficiency, and support the development of more proficient surgeons. This work demonstrates that fast, patient-specific VR simulations can provide interactive, case-based rehearsal of spinal decompression, with important implications for surgical education, pre-operative planning, and future competency-based training.
}




\maketitle

\section{Introduction}\label{sec1}

Simulation is commonly used in the education of surgical trainees as an important adjunct to didactic instruction and experience in the operating room. It supports learning of surgical processes, strategies, and technical skills and takes multiple forms, including physical models, scenario-based simulations, and virtual simulations\cite{b02}. Computer based simulation is widely applied in surgical education because it offers trainees an experiential learning environment in which practical skills can be built. Many aspects of surgery have a steep learning curve and simulation allows learning in a low-stakes environment\cite{b00,b01}. 

Virtual reality (VR) is an immersive technology that can provide a high-fidelity, cost-effective, and accessible option for surgical simulation\cite{b1,b4,b41}. VR-based surgical simulation can create an engaging and realistic training environment and has previously been shown to be beneficial in surgical education across specialties, including  orthopaedic, arthroscopic, ophthalmic, and neurosurgery disciplines \cite{b4}. It offers a safe and standardized method for training without the inherent risks to the operating room\cite{b00}. Evidence suggests that VR training improves cognitive and procedural performance, with measurable gains in speed and accuracy across a range of procedures\cite{b27}. More broadly, simulation allows trainees to practice skills without the risk of complications in clinical scenarios\cite{b28,b29}. With the transition toward competency-based curricula, the integration of VR simulation into traditional learning models may provide additional educational benefit\cite{b30}. 

The availability of VR training modules independent of patient, cadaver, or synthetic models allows a cost-effective, low-risk approach to surgical training with graded, unlimited practice opportunities \cite{Seymour2002VR,Zendejas2013Simulation}. With the growing limitations on operating room time and the potential for complications, VR-based simulation emerges as an effective experiential learning tool for surgical trainees. However, VR based training usually takes the form of standardized, off-the-shelf scenarios that may not directly link to a learner’s particular challenges.  Based on feedback from trainees and educators pre-made cases do not get ongoing use because they do not integrate into the competency-based curricula and the apprenticeship style learning \cite{Aggarwal2010Simulation}. The systems do not change with cases that the surgical trainees see in their training, creating a disjointed resource with limited relevance to the specific case at hand. In our experience, this contributes to underutilization of VR simulation within surgical training programs.  

Multimodal medical imaging, MRI and CT, is routinely used for treatment planning in spine surgery, primarily through inspection of two-dimensional slices. This requires surgeons to mentally reconstruct  a 3D representation of patient anatomy\cite{b43}, a cognitively demanding process that required extensive experience. Although 3D models, with or without VR integration, have been explored, their routine clinical use remains limited, particularly when multi-modal imaging visualization (such as in the spine) is required. 3D model use is limited because it requires significant time to create, technical expertise, and the use of multiple software packages\cite{b44}. Careful review of medical imaging is crucial for understanding how anatomy of the patient and any spinal pathology affect the surgical strategy. However, the use of 3D models in this review is currently too expensive and time-consuming to be performed routinely for surgical planning or education. 

The present investigation aims to develop an integrated framework for patient-specific 3D modelling, with the goal of allowing surgical practice in VR on cases scheduled for surgery. This work differs from previous VR based surgical education system because it builds patient specific models rather than using pre-made VR models, and by enabling interactive simulation of surgical tools rather than visualization alone.

This study focused on simulating spinal decompression procedures to treat spinal stenosis, a condition characterized by narrowing of the spinal canal, leading to compression of neural elements (spinal cord or nerves roots) \cite{b2}. This compression can result in debilitating pain, weakness, and numbness. Decompression surgery involves resection of tissues such as ligament, bone, tumour, or intervertebral discs to relieve compression of the neural elements. The success of such surgeries depends on an in-depth understanding of the patient-specific pathoanatomy, along with strategies to maximize patient benefit while minimizing risk.

The objectives of this study were to implement and evaluate: (1) automatic generation of 3D anatomical spine anatomical models from patient imaging, and (2) VR simulation of spinal decompression procedures using these models, including laminectomy, disc resection, and foraminotomy. The proposed methods were designed to be user-friendly, efficient, and accurate, facilitating the generation of high-fidelity, patient-specific models for interactive use. In terms of educational integration, simulation-based learning offers an interactive and effective approach to surgical training by enhancing engagement, spatial understanding, and procedural rehearsal.

VR simulations have been applied in many surgical disciplines, including general surgery, cardiac surgery, orthopaedic surgery, spine surgery, neurosurgery, and urology\cite{b4}. VR technology has proven particularly beneficial in medical education by providing an interactive and immersive training environment \cite{b41}.

\begin{figure*}[t]
    \centering
    \includegraphics[width=1\linewidth]{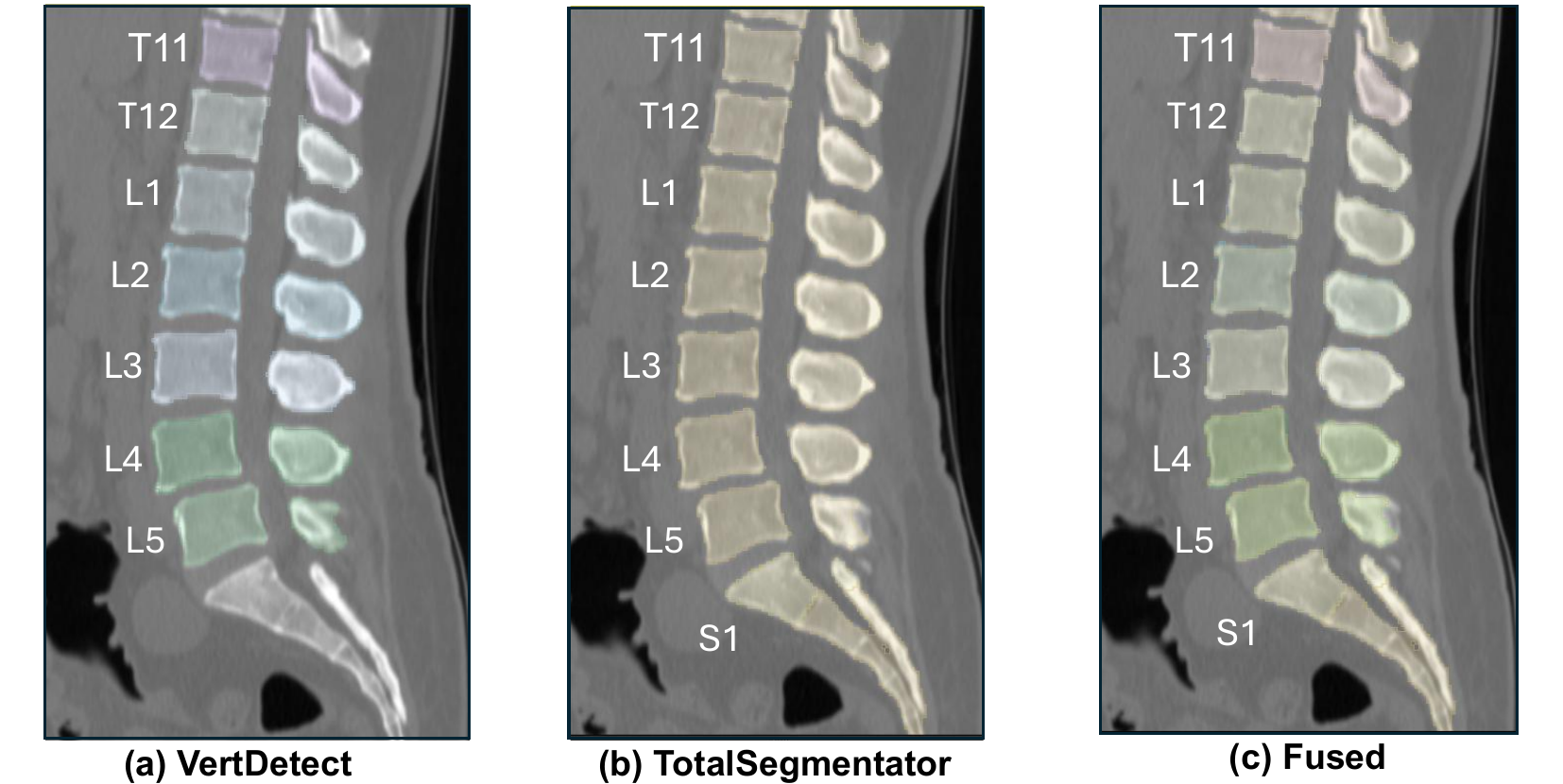}
    \caption{A sagittal plane of a 3D CT image of the lumbar spine is shown, overlaid with segmentation results and vertebral labels. The vertebrae are segmented (a) using VertDetect \cite{b12} and (b) using TotalSegmentator \cite{b13}, which includes segmentation of the sacral region. (c) A fused segmentation is generated by combining the outputs of VertDetect and Totalsegmentator to improve segmentation completeness and robustness.}
    \label{CT}
\end{figure*}

D. M. Croci et al. \cite{b6} introduced a 3D-VR tool named SpectoVR for visualization of medical imaging data, designed for preoperative planning in complex spine surgeries. The platform performs real-time volume rendering of DICOM data from CT, MRI, or combined modalities using ray casting within a virtual operating theatre environment. The authors reported that this platform is suitable, reliable and provides improved spatial understanding of a wide range of spine pathologies, including degenerative conditions, trauma, and tumours. However, SpetroVR is primarily limited to visualization. In contrast, the present investigation extends beyond visualization by incorporating interactive features, including virtual resection capabilities of anatomical structures. 

S. Zawy Alsofy et al. \cite{b7} evaluated a VR-based approach for surgical planning in patients with monosegmental, unilateral cervical foraminal stenosis (n=73). The study involved preoperative and postoperative CT scans, for the creation of 3D-VR representations, which were compared with conventional 2D slices of  CT slice based assessments for surgical planning, strategy and technique (ventral versus dorsal approach). Using the 3D Slicer, the authors quantified the smallest cross-sectional area of the intervertebral neuroforamen in the lateral resection region and surveyed the surgical strategy through questionnaire. Although, significant differences were observed between surgical approaches in terms of the foraminal area, no correlation was identified with clinical outcomes. This work demonstrates the utility of 3D-VR visualization in surgical planning; however, the approach focused on imaging assessment rather than interactive procedural simulation. 

Similarly, 3D models have been developed for preoperative planning in adolescent idiopathic scoliosis \cite{b8}. In a cohort of 60 patients undergoing posterior arthrodesis, preoperative planning using VR-based 3D anatomical models (n = 30) was compared with conventional CT-based visualization (n = 30). The results included operative time, blood loss, hospital stay, and surgeon satisfaction. VR-assisted planning was associated with reduced operative time and blood loss, as well as increased surgeon satisfaction. While these findings support the value of VR in surgical planning, the models were used primarily for visualization and did not incorporate patient-specific interactive simulation of surgical procedures.

A VR-based training simulator for the assessment of lateral lumbar approach was introduced by \cite{b10}, incorporating the 3D Systems Touch Haptic Device, Oculus Rift, and a cadaver spine model to improve the understanding of the surgical technique. The study demonstrated promising outcomes in improving surgical training for residents and less experienced surgeons. However, the platform did not incorporate patient-specific anatomical or pathological variability. The present investigation extends this line of work by creating patient-specific models directly from medical imaging, enabling simulation tailored to individual anatomy. This approach may support the adaptation of surgical strategies to patient-specific variations, thereby improving the educational relevance of simulation-based training.

D. Anthony \cite{b11} explored patient-specific VR technology for complex neurosurgical cases, developing 3D VR models for five adult patients undergoing procedures for spinal cord cavernoma, clinoidal meningioma, anaplastic oligodendroglioma, giant aneurysm, and arteriovenous malformation. These 360-degree VR models were generated by digitally rendering preoperative CT, MRI, or diffusion tensor imaging (DTI) using the Surgical Theater visualization platform. The study highlighted the potential of VR visualization for preoperative planning and surgical approach optimization in neurosurgery. However, the emphasis remained on immersive visualization and navigation rather than interactive procedural simulation.

Overall, the existing literature predominantly focuses on VR-based visualization of medical imaging for education and preoperative planning, typically relying on a single imaging modality. Prior investigations have not incorporated interactive surgical simulation features, such as tissue resection, that allow dynamic engagement with patient-specific anatomy. Furthermore, integration and fusion of multimodal imaging data, particularly combined CT and MRI for interactive simulation, remain limited. The present study addresses these gaps by developing a comprehensive 3D modelling framework that integrates multimodal imaging and enables interactive patient-specific surgical simulation within a VR environment. To our knowledge, this is the first integrated framework to enable fast, automated patient-specific spine models from CT and MRI data for fully interactive VR-based rehearsal of spinal decompression procedures. 

\begin{figure}[t!]
    \centering
    \includegraphics[width=1\linewidth]{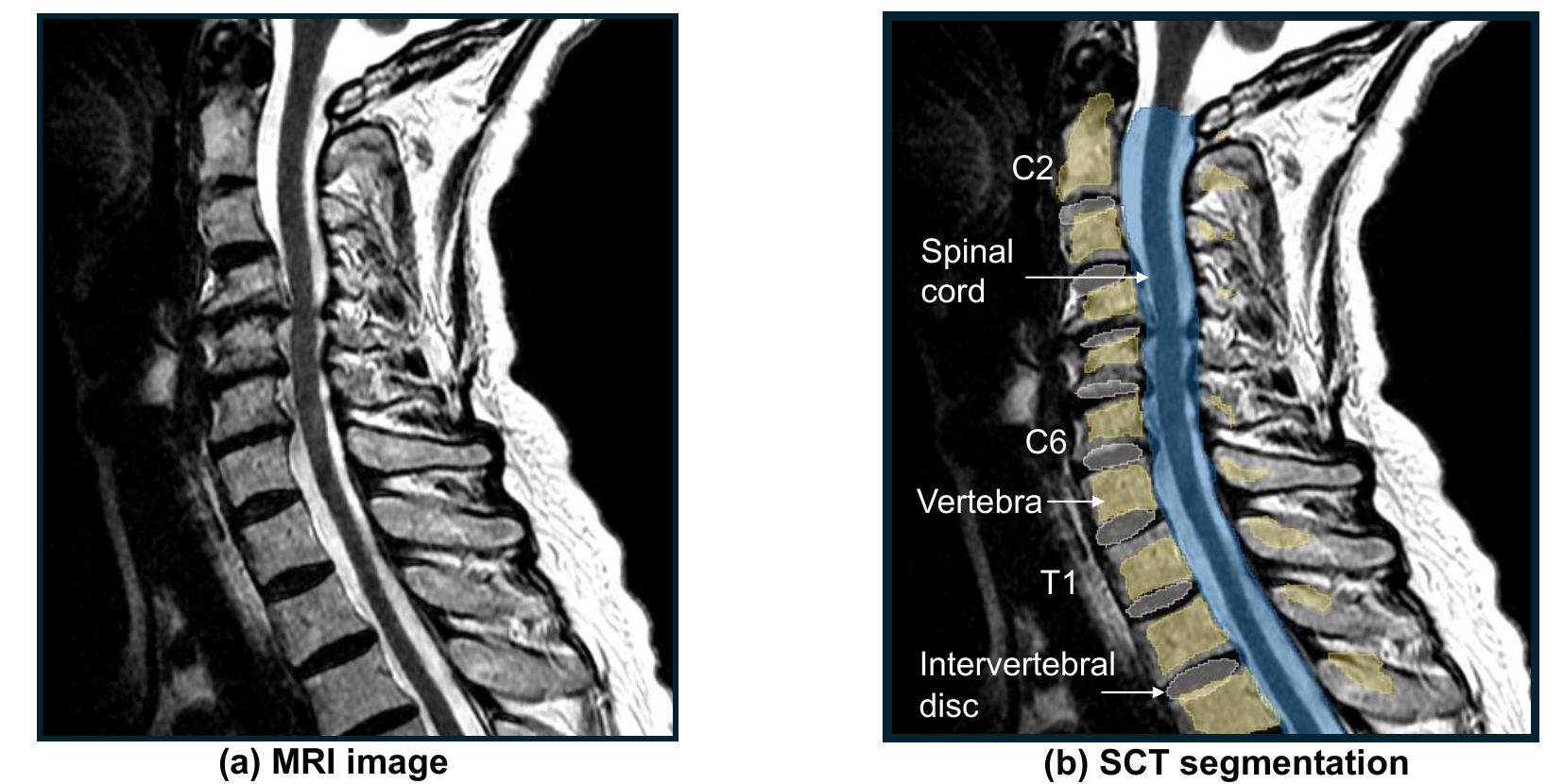}
    \caption{A sagittal plane of a 3D MRI image of the cervicothoracic spine. (a) MRI image. (b) Corresponding segmentation generated using the Spinal Cord Toolbox (SCT) \cite{b16}, demonstrating automatic identification of vertebral levels (e.g., C2, C6, T1) and segmentation of the vertebrae, intervertebral discs, and spinal cord.}
    \label{SCT}
\end{figure}


\section{Materials and Methods}
\subsection{Study Design and Data}

A semi-automated image analysis pipeline (Fig. \ref{pipeline}) was developed to utilize standard diagnostic medical imaging (CT and MRI) to create 3D models of the spine, enabling patient-specific simulation of spinal decompression surgery. The workflow included the detection and identification of vertebrae, the registration of multimodal data (MRI and CT), and the segmentation of neurological and musculoskeletal structures, including cerebrospinal fluid, spinal cord, nerve roots, bones, muscles, and intervertebral discs. CT imaging was used to access the vertebrae, whereas MRI provided superior soft tissue contrast for segmentation of neural elements and intervertebral discs. Advanced segmentation and registration methods using custom deep learning architectures and open-source tools, produced high-fidelity models integrated into a virtual reality application to simulate spinal decompression within a virtual operating room. The face validity of the simulation was assessed using structured feedback obtained from staff surgeons and trainees. 

This study was conducted as a retrospective analysis of clinical imaging data from 15 patients (8 lumbar, 3 cervical, 2 cervicothoracic, and 2 thoracic) who underwent spinal decompression procedures at Sunnybrook Health Sciences Centre. Institutional Research Ethics Board approval was obtained prior to data collection and analysis.

\subsection{Image Segmentation}
\subsubsection{Segmentation of vertebra in CT imaging}

A full 3D end-to-end vertebrae segmentation methodology (VertDetect) \cite{b12} was developed in our lab utilizing a Convolutional Neural Network (CNN) to predict vertebral level labels and segment individual vertebrae in CT imaging, as depicted in Fig.~\ref{CT}(a). Quantitative measures demonstrated high performance, with a DSC of 0.869 (95\% CI: 0.832, 0.891) in the VerSe 2020 public test sets. This work presents a faster, more accurate, and robust alternative for segmenting Vertebra in 3D CT images. This model has been integrated into the current pipeline to provide precise vertebral detection and segmentation. To enhance anatomical coverage, an additional deep learning segmentation model, TotalSegmentator\cite{b13} based on the nnU-Net architecture, was incorporated. TotalSegmentator segments 104 anatomic structures on CT images and reported a DSC of 0.943 (95\% CI: 0.938, 0.947) on a heterogeneous test set including significant abnormalities. In addition, this model enables sacral segmentation, a capability not available in the VertDetect framework as shown in Fig.~\ref{CT}(b). The vertebra segmentations of both models were fused  using a voxel-wise union operation to create a single, unified segmentation mask (Fig.~ \ref{CT}(c)), which was subsequently used for 3D bone modeling in VR simulation. 

\begin{figure}[t]
    \centering
    \includegraphics[width=1\linewidth]{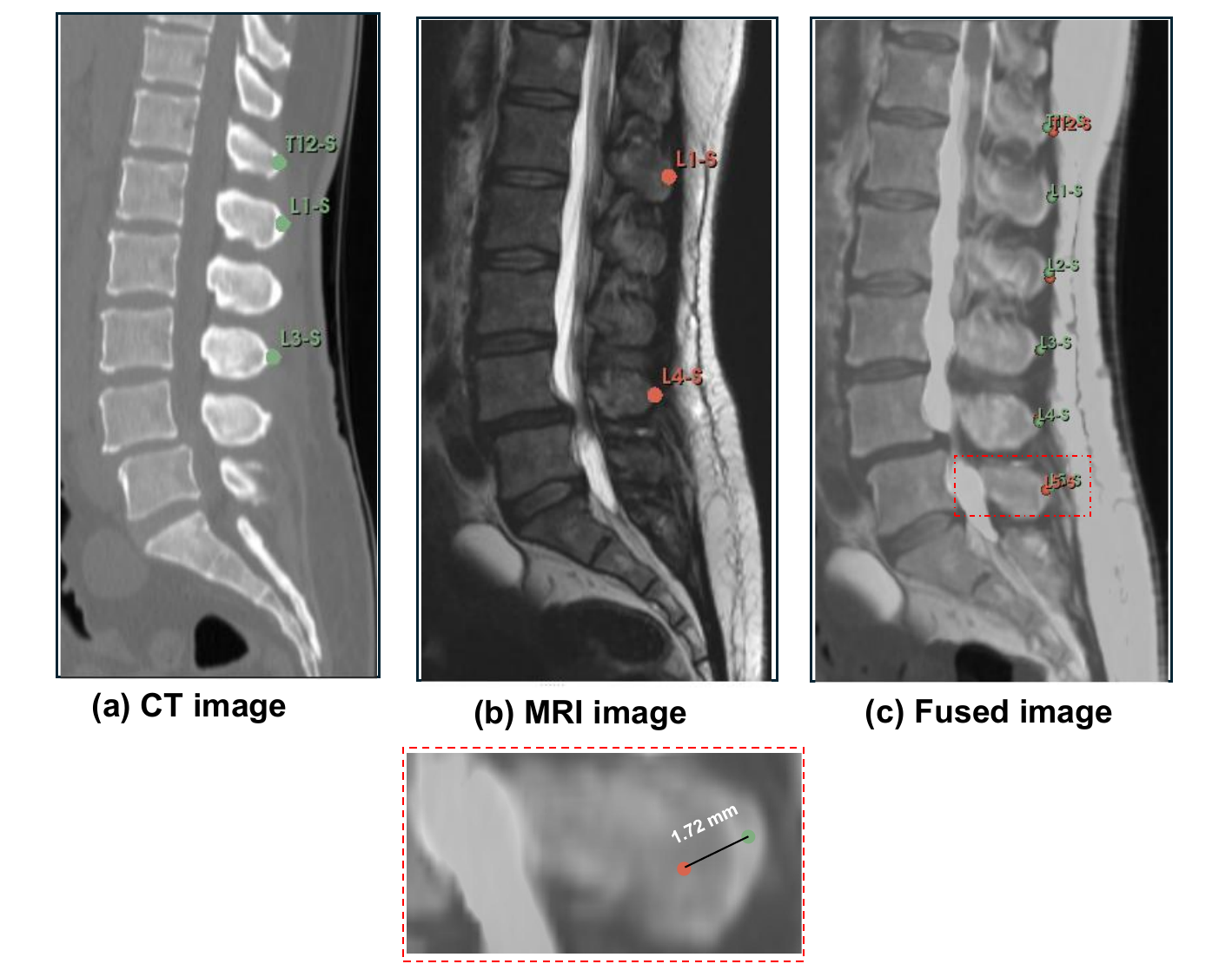}
    \caption{Multimodal CT-MRI registration of the lumbar spine. (a) CT image with manually placed vertebral fiducial landmarks (green). (b) MRI image with corresponding manually placed fiducial landmarks (red). (c) Fused CT-MRI image following landmark-based affine alignment and deformable registration. Landmarks were positioned at the posterior tip of the spinous process and transverse processes. The inset illustrates a representative total registration error (TRE) measured between corresponding landmarks located at the posterior tip of the spinous process of L5 vertebra.}
    \label{Registration}
\end{figure}

\subsubsection{Segmentation of vertebra and soft tissues in MRI imaging}

MRI images were segmented using two methods: (1) a custom trained U-Net based model for the lumbar spine, and (2) the Spinal Cord Toolbox (SCT) for the cervical and thoracic spine segmentation. The custom trained U-Net was trained using a publicly available, multi-center dataset for lumbar spine segmentation in MR images \cite{b14}. The dataset includes segmentations of vertebrae, intervertebral discs, and the neural elements. It consists of 218 patient studies with 447 sagittal T1 and T2 MRI series from 218 patients with a history of low back pain, including conditions such as disc herniation, spondylolisthesis, disc narrowing, disc bulging, Pfirrmann grade, upper and lower endplate changes, Schmorl’s nodes, and Modic changes. Forty-one of the images are T2 SPACE sequence images with near isotropic spatial resolution of 0.9 x 0.47 x 0.47 mm voxel size. T1 and T2 weighted sequences have voxel sizes ranging from 3.15 × 0.24 × 0.24 mm to 9.63 × 1.06 × 1.23 mm \cite{b14}. 

The nnU-Net framework \cite{b15} was used to train a deep learning model using the 3D full-resolution configuration. The nnU-Net model is based on the U-Net architecture, a convolutional neural network (CNN) designed for image segmentation. It is a widely used and powerful semantic segmentation method that automatically adapts network architecture and preprocessing to a given dataset. An example of a lumbar spine segmentation generated by the trained nnU-Net model, including the vertebrae, intervertebral discs, and the spinal canal is shown in Fig.~\ref{pipeline}(d). The ligamentum flavum and nerve roots were manually contoured using 3D Slicer (version 5.6.2) to complete 3D anatomical model, as illustrated in Fig.~\ref{pipeline}(d)(f).

 MRI of the cervical and thoracic spine were segmented using SCT\cite{b16}. SCT is a free and open-source platform with a set of command-line tools dedicated to the processing and analysis of spinal cord MRI data, which can also be extended to the vertebra segmentation. The pipeline begins with spinal cord segmentation in anatomical MRI images. Previous investigations reported by SCT achieved a DSC of 0.95 ± 0.03 using CNN with 3D convolutions and 0.89 ± 0.16 using the PropSeg algorithm for spinal cord segmentation \cite{b17}. Spinal cord vertebral levels or intervertebral discs are automatically identified in images containing the C2-C3 disc. If automated identification fails or the MRI does not include the C2-C3 disc, manual labeling of vertebrae is performed.

To segment nerve roots, intervertebral discs and ligamentum flavum, manual contouring was performed using 3D Slicer (version 5.6.2) on the PAM50 template. The PAM50 template is a structural template for spinal cord imaging derived from 50 healthy subjects \cite{b17-1}. It spans vertebral levels C1 to L2. Patient MRI images that have been segmented and labeled are then registered with the PAM50 template to identify all neurological (CSF, spinal cord, nerve roots) and musculoskeletal structures (bone, ligamentum flavum, intervertebral discs). Segmentation using SCT is illustrated in Fig.~\ref{SCT}.

The deep learning algorithm training was done using Narval (Nvidia DGX A100 (40 GB memory)), part of the Niagara resources, managed by the Digital Research Aliance of Canada and the High Performance Computing (HPC) facilities at Sunnybrook Research Computing (Nvidia DGX A100 (80 GB memory)) funded by Canada foundation for innovation (40206), and the Ontario Research Fund.

\subsection{CT and MRI image registration}

The third stage of the workflow involved registration of CT and MRI images. Registration was performed in two stages: landmark-based affine alignment followed by deformable registration.

3D landmark locations were automatically identified using the segmented vertebrae in both MRI and CT images. The centroid of each vertebrae was computed at each spinal level and utilized to allow spatial alignment between CT and MRI\cite{b20,b21}. Based on these landmarks, a similarity transformation matrix was estimated using the Fiducial Registration Wizard module in 3D Slicer\cite{b3}. First, the centroids of the vertebrae in the MRI (moving image) and CT (fixed image) were calculated and aligned by subtracting their respective mean positions to remove translation offset. A cross-covariance matrix was then constructed to describe the relationship between the two point sets. Singular Value Decomposition (SVD) was then used to compute the optimal rotation matrix. A uniform scaling factor was calculated to account for proportional differences between CT and MRI. Finally, a translation vector was computed by aligning the centroids after rotation and scaling. These components were combined to form a similarity transformation matrix, which was subsequently applied to align the moving MRI image to the fixed CT image.

\begin{figure*}[t]
\centering
\includegraphics[width=1\linewidth]{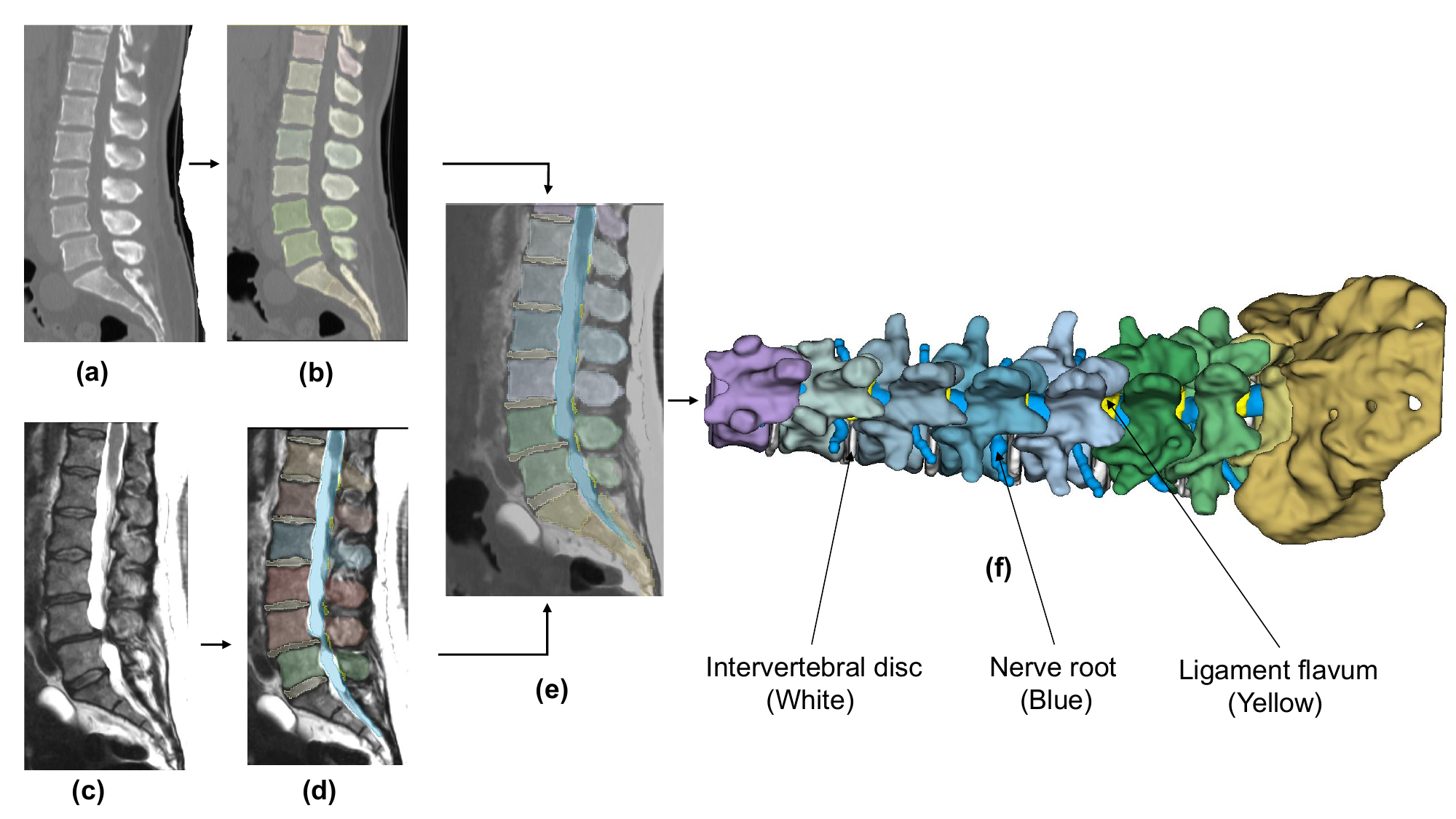}
\caption{Overview of the proposed workflow for creating patient-specific 3D spine models for virtual reality simulation of spine procedures.(a) CT image. (b) CT segmentation. (c) MRI image. (d) MRI segmentation. (e) Multimodal registration (affine and deformable). (f) Reconstruction of fused segmentations into a patient-specific 3D model.}
\label{pipeline}
\end{figure*}

Following affine alignment, deformable image registration was performed (Fig.~\ref{pipeline}(e)). Soft tissues of the spine, including the spinal cord, intervertebral discs, nerves roots, and ligamentum flavum, may exhibit deformation between modalities. A perturbed continuous optimization approach developed in our laboratory \cite{b22} was used to deform the MRI image to align with the reference CT image.

This deformable registration framework optimizes a displacement field (\(D\)) to align a moving image (\(m\)) with a fixed image (\(f\)) using continuous optimization. The ADAM optimizer was employed to upgrade gradients with momentum, reducing oscillations, and improving convergence stability. Image similarity was evaluated using Modality Independent Neighbourhood Descriptor (MIND) features to enable robust multiomodal alignment. The similarity metric was defined as:

\[
S = \frac{1}{N} \sum \left(F_f - F_m\right)^2
\]

where:
\begin{itemize}
    \item \(N\): Total number of voxels in the image,
    \item \(F_f\): MIND features of the fixed image,
    \item \(F_m\): MIND features of the warped moving image.
\end{itemize}

Minimizing of \(S\) promotes alignment of the two images in the feature space. To enforce smooth deformations, a regularization term was introduced to penalize large gradients in the displacement field:

\[
R = \frac{1}{N} \sum |\nabla D|
\]
h
where \(|\nabla D|\) represents the gradient magnitude of the displacement field. This regularization promotes spatial smoothness and prevents unrealistic deformations.

The learning rate (\(\eta\)) was adapted over 250 iterations (\(s\)) using a Piecewise Decay (PWD) schedule, defined as:

\[
\eta(s) =
\begin{cases}
15 & 0 \leq s < 70, \\
7 \cdot \cos\left(\frac{2\pi}{200} \cdot (s - 70)\right) + 8 & 70 \leq s < 180, \\
-\frac{1.209}{70} \cdot (s - 180) + 1.343 & 180 \leq s \leq 250.
\end{cases}
\]

The PWD schedule enables efficient optimization by starting with a high learning rate (\(15\)) for rapid convergence in the early iterations, transitioning to a cosine decay phase for refinement, and ending with a linear decay phase to achieve sub-voxel accuracy.

The continuous optimization is a rapid and precise method inspired by convex ADAM \cite{b23} and has been previously evaluated on the NLST and OASIS datasets, demonstrated robust performance in estimating large deformations.

Registration accuracy was assessed using Total Registration Error (TRE) calculated after completion of deformable registration from manually placed anatomical landmarks identified in both MRI and CT images. Three anatomical landmarks were selected at each evaluated vertebral level: (1) the posterior tip of the spinous process, and (2-3) the mid-aspect of the left and right transverse processes. Landmarks were initially placed  on unregistered MRI and CT images in the sagittal view and subsequently refined using coronal and axial views in 3D Slicer as shown in Fig.~\ref{Registration}.

TRE was defined as the Euclidean distance between corresponding landmark coordinates in the registered MRI and CT volumes. TRE values were computed per landmark and subsequently averaged at three levels: (1) per landmark, (2) per vertebra, and (3) per patient. The final reported TRE represents the mean ± standard deviation across patients.

Landmarks were chosen in regions where the spatial relationship between bony and adjacent soft tissue structures is clearly identifiable and consistent with previous literature \cite{b24}\cite{b25}. These  anatomical structures are rigid, well-defined, and visible in both imaging modalities. The final landmark placement was  independently confirmed by two orthopaedic surgeons to ensure accuracy and reproducibility.

\subsection{3D Model Construction and VR Simulation}

3D models (Fig.~\ref{pipeline}(f)) were exported in DICOM Seg format using 3D Slicer (version 5.6.2). The segmented models and associated imaging data were then imported into a custom spine surgery simulation module integrated within the SieVRt VR DICOM viewing platform (Luxsonic Technologies, Saskatoon, Canada). 

The spine surgical simulation module enabled users to perform a simulated spinal decompression procedure using patient-specific models within a virtual operating room environment. This module allowed users to simulate spinal decompression. Trainees and faculty can use this module and its functionality for educational purposes and treatment planning.   

3D virtual patient specific models were presented as they would be when prepared for surgery, in the prone position on a surgical table. Surgical exposures were automatically generated, with manual adjustments possible.  Surgical exposures allowed visualization of the spinal levels being decompressed. Users could collaboratively perform surgery, discuss the surgery with another user, switch between performing or observing the surgery.  Users could choose from surgical instruments (burr, kerrison punch, woodson, rongeur) needed to perform the procedure from a virtual surgical tray. Real-time mesh modification was implemented to dynamically update anatomical structures during simulated tissue removal. 

The simulation allows review of medical imaging within the simulated OR using dedicated panels, allowing reference to diagnostic imaging while practising the surgery and better linking the anatomical structures and pathology to the 3D representation. In addition, the system provided auditory feedback through a burring sound that changes pitch based on what it is in contact with, further immersing the user in the surgical environment. To give users further information about proximity to the neural structures a visual and auditory alarm system was implemented. No haptic feedback was incorporated in the current implementation.

For post-procedural assessment, users could isolate spine anatomy from the surrounding virtual patient body to allow detailed and improved three-dimensional visualization.  Anatomical structures could be rotated and magnified to evaluate the adequacy of decompression and surgical approach. 

\subsection{Evaluation Protocol}

\subsubsection{Quantitative Evaluation}

Performance of the image analysis pipeline was evaluated using the Dice Similarity Coefficient (DSC) to assess volumetric agreement between automated  and reference segmentations. Registration accuracy was quantified using Target Registration Error (TRE), computed after the deformable registration as described in the Registration section. Computational performance was assessed by recording processing time for segmentation, deformable registration, and total end-to-end model generation for each case. All experiments were conducted on a Windows workstation equipped with an Intel i9-14900KF CPU (3.20 GHz), 32 GB RAM, and an NVIDIA RTX 4080 SUPER GPU.

\subsubsection{Qualitative Evaluation}

Semi-structured interviews were conducted with two staff surgeons and three trainees to assess the qualitative value of VR-based surgical simulation for spinal decompression procedures. The interviews aimed to qualitatively evaluate impact on medical practice, usability, learner confidence, and engagement\cite{b26}.

\paragraph{Medical Practice}
The following domains were included in the interview framework:

\begin{itemize}
    \item \textbf{Prior Knowledge} -– Participants' familiarity with spinal decompression procedures and their ability to interpret patient-specific anatomical structures.
    \item \textbf{Resectability} -– Confidence in evaluating surgical feasibility and perceived accuracy of the VR model in representing real patient anatomy.
    \item \textbf{Recall} -– Retention of 3D model’s spatial information during surgery and it's effect on preparedness.
    \item \textbf{Added Value} -– Perceived benefits of VR for spatial understanding, identification anatomical variations, and  surgical decision-making.
    \item \textbf{Uncertainty in Imaging} -- Perceived mismatches between MRI images and the VR model.
    \item \textbf{Workflow Integration} -- Potential incorporation of VR into existing clinical workflows, including interdisciplinary discussion, education, and surgical planning
\end{itemize}

\paragraph{Usability}
The following usability domains were explored during the interviews:

\begin{itemize}
    \item \textbf{Ease of Use} -- Perceived intuitiveness and user-friendliness of the VR system..
    \item \textbf{Spatial Understanding} -- Perceived improvement in three-dimensional spatial awareness during interaction with VR models.
    \item \textbf{Given Functionalities} -- Perceived usefulness of features such as resection , MRI integration, and stepwise planning.
    \item \textbf{Usability Issues} -- Reported challenges, including headset discomfort and instrument precision.
    \item \textbf{Requested Features} -- Suggested enhancements, such as measurement tools and visibility controls.
    \item \textbf{Multiuser Support} -- Perceived value of collaborative VR interaction for education and surgical planning.
\end{itemize}

\section{Results}


\renewcommand{\arraystretch}{1.3}
\begin{table*}[t]
    \centering
    \caption{Segmentation accuracy (DSC), registration accuracy (TRE), and computational time for patient-specific 3D model generation. Values are reported as mean $\pm$ standard deviation across patients.}
    \begin{tabular}{ccc}
        \toprule
        \textbf{Spinal Component} & \textbf{Performance (DSC / TRE)} & \textbf{Compute Time (s)} \\
        \midrule
        Vertebral Bone & $0.95 \pm 0.03$ & 72.60 \\
        Intervertebral Discs & $0.87 \pm 0.04$ & 31.20 \\
        Neural Elements & $0.92 \pm 0.01$ & 31.20 \\
        \midrule
        Overall TRE (mm) & $1.73 \pm 0.42$ & -- \\
        \midrule
        Registration Time (s)  & --  &  20.0 \\
        Total Model Generation Time & -- & $\sim$155\ \text{(2.58 min)} \\
        \bottomrule
    \end{tabular}
    \label{results}
\end{table*}

\begin{figure*}[!t]
\centering
\includegraphics[width=1 \linewidth]{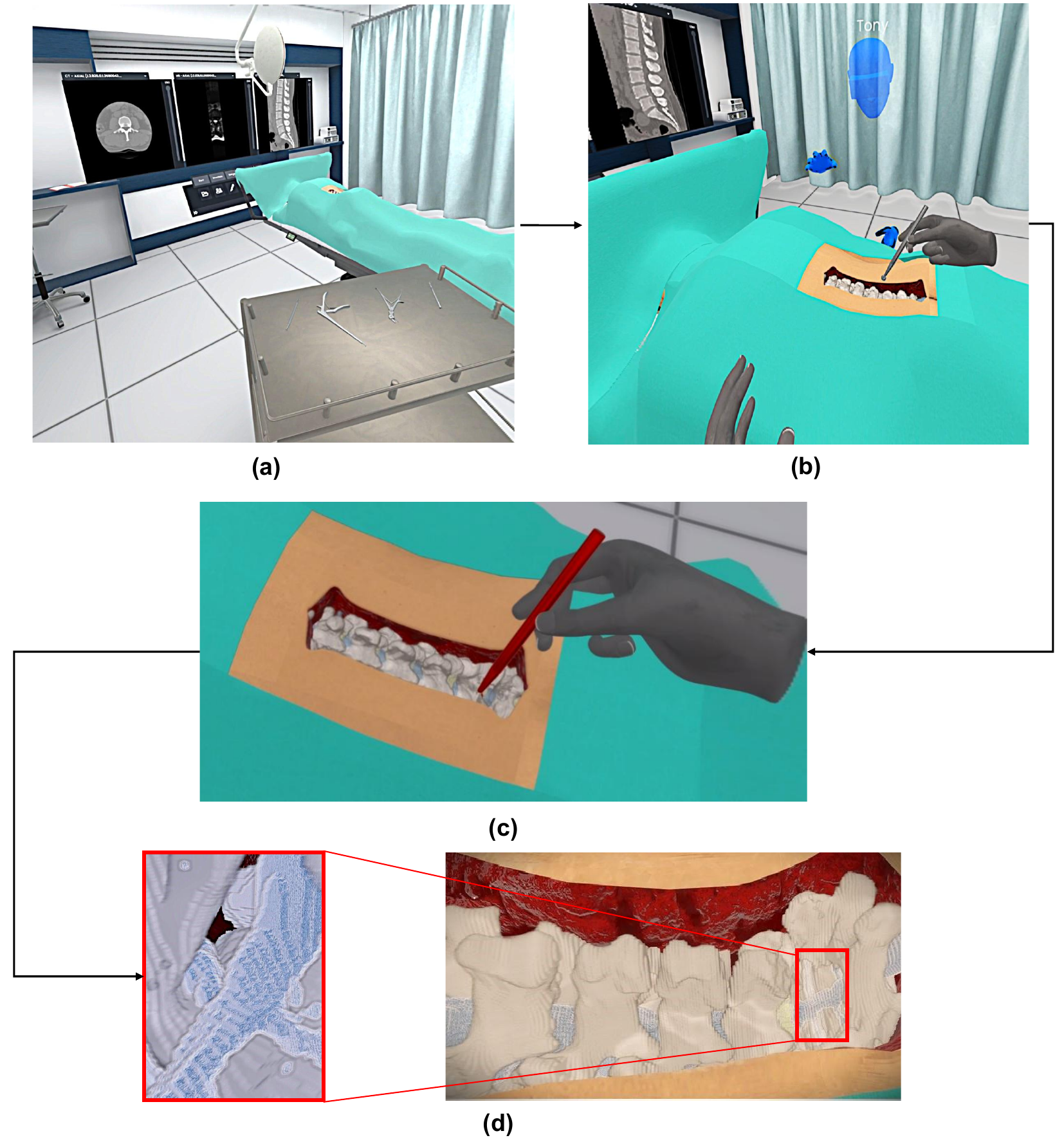}
\caption{
Virtual reality simulation of patient-specific spinal decompression within the SieVRt platform. (a) Virtual operating room environment with integrated multimodal imaging panels and surgical instruments.
(b) Collaborative simulation mode with automated surgical exposure, illustrating one user performing decompression with a virtual burr while a second user observes. 
(c) Simulated intraoperative decompression using a virtual burr; red highlighting indicates proximity to neural structures. 
(d) Magnified view of the  decompression region demonstrating visualization of the dura and nerve roots at L4–L5 level.
}
\label{sievrt}
\end{figure*}
\subsection{Quantitative Evaluation}

High-fidelity, patient-specific 3D models were created successfully for all 15 patients included in the study. The mean total model creation time was approximately 155 seconds (2.5 minutes) per case. vertebral bone segmentation achieved a mean DSC of 0.95 ± 0.03. Intervertebral disc segmentation achieved a mean DSC of 0.87 ± 0.04, while neural element segmentation achieved a mean DSC of 0.92 ± 0.01. Following deformable registration, the mean Target Registration Error (TRE) was 1.73 ± 0.42 mm across patients. Detailed segmentation accuracy, registration performance, and computational time metrics are summarized in Table~\ref{results}

\subsection{Qualitative Evaluation}
\subsubsection{Medical Practice}
Participants highlighted the added value of VR simulation in improving their understanding of spinal decompression procedures, particularly in visualizing critical anatomical structures and practicing procedural steps. The ability to combine anatomical knowledge with patient-specific medical imaging was regarded beneficial.

\begin{itemize}
    \item \textbf{Prior Knowledge} –- Trainees reported that merging anatomical references with patient-specific imaging improved their ability to visualize the neural foramen and understand patient-specific anatomical variation, particularly in complex cases. 
    \item \textbf{Resectability} –- The ability to see into the foramen and visualize the decompression required for nerve root treatment, particularly in the lateral recess, was identified as particularly valuable. Participants noted that the VR system provided a visual experience similar to that of a surgical microscope, which enhanced confidence in the evaluation of surgical feasibility.
    \item \textbf{Recall} -– Trainees appreciated the ability to go through procedures step by step, reinforcing their understanding of the process and improving their ability to recall spatial information. This repetition also helped solidify their knowledge related to anatomical structures and their relationships.
    \item \textbf{Added Value} –- The 3D modelling of pathology was recognized as helpful for understanding specific cases and determining the appropriate surgical approach. Trainees felt that practising with actual cases helped them prepare more effectively for real procedures, improving both decision-making and procedural confidence.
    \item \textbf{Uncertainty in Imaging} –- While no direct mention was made on uncertainty in imaging, some trainees acknowledged the challenge of aligning medical imaging with the VR model. However, merging anatomical text with patient-specific imaging helped clarify doubts, and the accuracy of the models seemed to resolve most issues.
    \item \textbf{Workflow Integration} –- The VR simulation was considered as a useful tool for practising real cases, which trainees believed would be helpful for increasing their surgical breadth and experience. It was noted that the platform could expedite learning by simulating OR environments.
\end{itemize}

\subsubsection{Usability and System Evaluation}
Participants reported that the system usability and its ability to improve spatial understanding were key advantages, although some areas for improvement were identified.

\begin{itemize}
    \item \textbf{Ease of Use} -- The VR system allowed them to work through the decompression procedure step by step, enhancing their ability to visualize and understand the decompression process. Some mentioned the potential usefulness of features such as haptic feedback or vibration to make the experience more immersive and realistic, potentially improving the learning experience.
    \item \textbf{Spatial Understanding} –- The VR system was particularly praised for improving three-dimensional spatial awareness. Participants noted that visualization of the lateral recess and nerve root decompression was clearer compared with conventional imaging review. The ability to inspect anatomy dynamically supported improved comprehension of decompression requirements in both the foramen and lateral recess.
    \item \textbf{Given Functionalities} –- While the VR system’s features were highly appreciated, trainees requested additional magnification for better visualization, particularly in areas that require more detailed observation, such as complex anatomical structures near the nerve roots. This enhancement could further improve the clarity and precision needed for successful surgery.
    \item \textbf{Usability Issues} –- No significant usability issues were reported, but the need for more magnification and possible improvements in feedback mechanisms (such as vibration) were mentioned. Trainees also expressed that some adjustments to the interface could improve the overall user experience.
    \item \textbf{Requested Features} –- Trainees expressed interest in additional magnification capabilities, which would further enhance the clarity of intricate details during the simulation. Furthermore, the ability to control visibility of specific anatomical structures was suggested as a valuable feature for detailed analysis.
    \item \textbf{Multiuser Support} –- The VR platform was believed to expedite hands-on learning by providing simulated operating rooms, which could potentially increase case volume and support team collaboration in training. Trainees suggested that multiuser support could facilitate collaborative learning and teamwork, further enhancing the educational experience.
\end{itemize}

\section{Discussion}

This study presents a fast and automated framework for creating patient-specific 3D spine models from multimodal CT and MRI data for use in immersive virtual reality simulation. The pipeline achieved high segmentation accuracy, low registration error (TRE 1.73 ± 0.42 mm), and rapid model generation (approximately 2.5 minutes per case), supporting the feasibility of integrating patient-specific simulation into educational and preoperative workflows. 

Most existing VR applications in surgical training and planning focus primarily on 3D visualization using volume rendering, often derived from a single imaging modality. Croci et al. \cite{b6} developed SpectoVR for spine visualization using real-time volume rendering, but the system did not include interactive procedural simulation such as virtual resection. Similarly, Zawy Alsofy et al. \cite{b7} applied 3D-VR based volume rendering for cervical foraminal stenosis to support surgical planning; however, the platform emphasized visualization and measurement rather than procedural interaction. Studies investigating VR in scoliosis planning \cite{b8} and lumbar surgery simulators incorporating cadaver spine models \cite{b10} demonstrated educational benefits, including reduced operative time and improved surgeon satisfaction. However, these systems did not incorporate automated creation of multimodal patient-specific models within an interactive simulation environment.

Anthony \cite{b11} explored VR models for neurosurgical planning, highlighting the value of immersive visualization for complex cases. While this work demonstrated the clinical utility of VR for anatomical comprehension, the emphasis remained on visualization rather than interactive surgical rehearsal. In contrast, the present work extends beyond visualization, allowing real-time simulation and interaction within a virtual environment, supporting case-based procedural rehearsal for trainee education. 

Our previous work \cite{b5} demonstrated the educational feasibility of VR simulation for spinal decompression but relied on manual segmentation and registration processes. Although effective, that approach was time-intensive and limited scalability. Feedback from trainees and faculty highlighted the need for automation to support routine integration into educational workflows. The current framework addresses this limitation by streamlining segmentation and multimodal registration, substantially reducing model preparation time and improving practical feasibility.

Qualitative feedback from staff surgeons and trainees indicated positive perceptions, particularly regarding enhanced understanding of spinal stenosis and pathoanatomy. The immersive experience was perceived to improve comprehension of complex surgical concepts and increase procedural confidence among trainees. The ability to repeatedly simulate surgical scenarios in a risk-free virtual environment can complement traditional training methods, which are often limited by cadaver availability and restricted operating room time.

The integration of patient-specific 3D models into VR-based simulations has important implications for surgical education and preoperative planning. Traditional surgical training relies on cadaveric dissections, textbooks, and 2D imaging review, which may not provide the depth of spatial understanding required for complex procedures such as spinal decompression. The present system provides an interactive environment in which trainees can rehearse procedures using patient-specific models derived from clinical imaging.

An advantage of this approach is the ability to simulate various surgical scenarios repeatedly, allowing trainees to build muscle memory and improve decision-making skills before performing real procedures. Unlike conventional training methods constrained by physical resources and access to the operating room, VR-based simulation offers a scalable alternative that supports consistent and reproducible learning experiences.

Furthermore, this framework may also contribute to preoperative planning. By allowing surgeons to interact with patient-specific anatomy in a virtual environment, they can anticipate potential surgical challenges, optimize their approach, and minimize intraoperative risks. The ability to visualize soft tissues from MRI alongside bony structures from CT data offers a more comprehensive understanding of the patient’s anatomy. Although the impact on clinical outcome was not assessed in this study, this multimodal integration may support improved surgical planning, particularly in complex or anatomically variant cases.

In addition, the VR system facilitates collaborative learning and interdisciplinary discussions. Faculty and trainees can engage in case-based discussions within the virtual environment, assessing different surgical approaches and receiving expert feedback in real-time. This fosters a more interactive and engaging educational experience, bridging the gap between theoretical knowledge and practical application.

Despite these potential benefits, successful adoption of this technology requires structured integration into surgical training curricula. Future work should focus on developing standardized training modules, assessment metrics, and competency-based evaluation systems to quantify the impact of VR-based learning. 

The quantitative evaluation was conducted on a relatively small sample (n=15) from a single institution, which may limit generalizability. Although segmentation accuracy and registration performance were favorable, further validation across larger and more diverse datasets is required. In addition, certain structures, including nerve roots and ligamentum flavum, required partial manual segmentation, indicating that complete automation has not yet been achieved.

Future work will focus on expanding automation of soft tissue segmentation (nerve roots and ligamentum flavum), and validating the framework in larger multicenter cohorts. Incorporation of objective educational and clinical performance metrics will be essential to evaluate the impact of the system beyond technical feasibility. Prospective studies examining workflow integration and influence on surgical planning decisions will further clarify clinical relevance.

Patient-specific simulation was designed to support surgical education and treatment planning, allowing trainees and faculty to practice procedures and understand the impact of pathoanatomy on the cases they manage. Furthermore, the simulations may facilitate discussions in which trainees and faculty interact with patient-specific geometries, offering a structured approach for receiving feedback and guidance on surgical strategy and performance.

\section{Conclusion}

In conclusion, our findings demonstrate that patient-specific 3D models integraded into VR simulations can support surgical training and preoperative planning by providing an interactive, immersive learning environment that enhances anatomical understanding, improves surgical confidence, and enables risk-free procedural rehearsal. Additionally, VR-based simulations may serve as a valuable tool for patient education, improving communication of surgical procedures and expected outcomes. Continued improvements in anatomical visualization, registration accuracy, and automation will further strengthen the clinical and educational value of this system. Ongoing advancements in segmentation and simulation design will be important for improving scalability and integration into surgical curricula.

\section*{Acknowledgements}

We sincerely thank all staff surgeons and trainees who participated in this study and provided valuable feedback on the virtual reality simulations; their insights were instrumental to this investigation.

We extend our appreciation to Normand Robert for his assistance in retrieving DICOM images from the clinical server, as well as for data management and secure handling of patient information. We also thank Geoffrey Klein for his expertise in CT vertebral segmentation and Teodora Vujovic for her significant contributions to the deformable registration work. Additionally, we are grateful to Cari Wyne for her valuable feedback.

We acknowledge Luxsonic Technologies for their collaboration in the development of the virtual reality platform.

\section*{Declarations}

\subsection*{Funding}

This work was supported by the Feldberg Chair for Spinal Research and INOVAIT Canada.

\subsection*{Conflict of Interest}

The authors declare that they have no competing interests related to this study. Luxsonic Technologies collaborated in the development of the virtual reality platform; however, the company had no role in the study design, data collection, data analysis, interpretation of results, or preparation of the manuscript.

\subsection*{Ethics Approval and Consent to Participate}

This study was approved by the Sunnybrook Health Sciences Centre Research Ethics Board. All procedures were conducted in accordance with institutional ethical standards. For the retrospective imaging component of this study, the requirement for patient informed consent was waived by the Research Ethics Board due to the use of de-identified data. Informed consent was obtained from all surgeons and trainees who participated in the semi-structured interviews.

\subsection*{Consent for Publication}

Not applicable. This manuscript does not contain any identifiable individual person’s data, images, or videos.

\subsection*{Data Availability}

The datasets generated and/or analyzed during the current study are not publicly available due to institutional privacy regulations but are available from the corresponding author upon reasonable request and with appropriate institutional approval.

\subsection*{Code Availability}

The code used to develop the image analysis pipeline and simulation framework is not publicly available due to institutional and collaborative agreements but may be made available from the corresponding author upon reasonable request and with appropriate institutional approval.

\subsection*{Author Contributions}

Raj Kumar Ranabhat contributed to study design, development of the image analysis pipeline, implementation, data analysis, and drafting of the manuscript. Tayler D. Ross contributed to clinical evaluation, qualitative analysis, and critical review of the manuscript. Tony Jiao contributed to manual segmentations and coordination of demonstrations and questionnaire administration with participants. Jeremie Larouche and Joel Finkelstein contributed to clinical oversight and manuscript revision. Michael Hardisty supervised the project, contributed to study design, data analysis and critically revised the manuscript. All authors read and approved the final manuscript.



\bibliography{sn-bibliography}

\end{document}